\documentclass[10pt,twocolumn,letterpaper]{article}
 
\usepackage{cvpr}              
\usepackage{pifont}

\usepackage{marvosym}

\definecolor{cvprblue}{rgb}{0.21,0.49,0.74}
\usepackage[pagebackref,breaklinks,colorlinks,allcolors=cvprblue]{hyperref}
\usepackage{adjustbox}
\def\paperID{} 
\def\confName{CVPR}
\def\confYear{2026}
\usepackage{multirow}
\usepackage{xcolor}
\usepackage[dvipsnames,svgnames]{xcolor}

\title{ReAlign: Generalizable Image Forgery Detection via Reasoning-Aligned Representation}

\author{
Qing Huang\textsuperscript{1}, Zhipei Xu\textsuperscript{2}, Xuanyu Zhang\textsuperscript{2}, Xiangyu Yu\textsuperscript{1,3 \Letter}, Jian Zhang\textsuperscript{2,4}\\
\textsuperscript{1} School of Future Technology, South China University of Technology \\
\textsuperscript{2} School of Electronic and Computer Engineering, Peking University \\
\textsuperscript{3} School of Electronic and Information Engineering, South China University of Technology \\
\textsuperscript{4} Guangdong Provincial Key Laboratory of Ultra High Definition Immersive Media Technology, \\ Shenzhen Graduate School, Peking University \\ 
\vspace{-30pt}
}

\begin{document}
\maketitle
\begin{abstract}
The rise of AI-generated images (AIGIs) poses growing challenges for digital authenticity, prompting the need for efficient, generalizable image forgery detection systems. Existing methods, whether non-LLM-based or LLM-based, exhibit distinct advantages and limitations. While non-LLM-based models offer efficient low-level artifact detection, they often lack semantic understanding. Conversely, LLM-based methods provide strong semantic reasoning and explainability but are computationally intensive and less sensitive to subtle visual artifacts. Moreover, the true contribution of explanatory reasoning texts to forgery detection performance remains unclear. In this work, we investigate the intrinsic value and potential of LLM-generated reasoning texts, considering it a source of generalization and semantic-error sensitivity. Based on these findings, we propose ReAlign, a novel framework that distills high-quality reasoning texts generated by a GRPO-optimized LLM into a lightweight AIGI detector via contrastive learning. ReAlign effectively inherits the generalization ability and semantic sensitivity capability of reasoning textual representations, while remaining efficient and lightweight for deployment. Moreover, ReAlign adopts a tailored joint optimization strategy that integrates contrastive loss for image-text alignment and classification loss for accurate forgery discrimination. Experimental results on AIGCDetectBenchmark, AIGI-Holmes, and our newly constructed UltraSynth-10k demonstrate that ReAlign consistently outperforms existing state-of-the-art detectors in both accuracy and generalization, particularly when facing complex, high-fidelity forgeries from modern generative models.
\vspace{-20pt}
\renewcommand\thefootnote{\relax} \footnote{\Letter: Corresponding author.}
\end{abstract}    
\section{Introduction}
\label{sec:intro}

With the rapid development of deep learning~\cite{yu2024identifying,shen2025ai,li2023ultrare,li2025multi} and generative technologies~\cite{wu2025qwen,zhang2025molebridge,zhang2025strfilter}, AI-generated images (AIGIs) have become increasingly widespread, significantly lowering the barrier to producing highly realistic images. However, their misuse poses security and ethical risks such as privacy breaches and the spread of misinformation, undermining public trust in digital media. Therefore, there is an urgent need to develop an efficient, reliable detection system to address the potential threats posed by AIGIs.

\begin{figure}[t]
	\centering
	\includegraphics[width=1.0\linewidth]{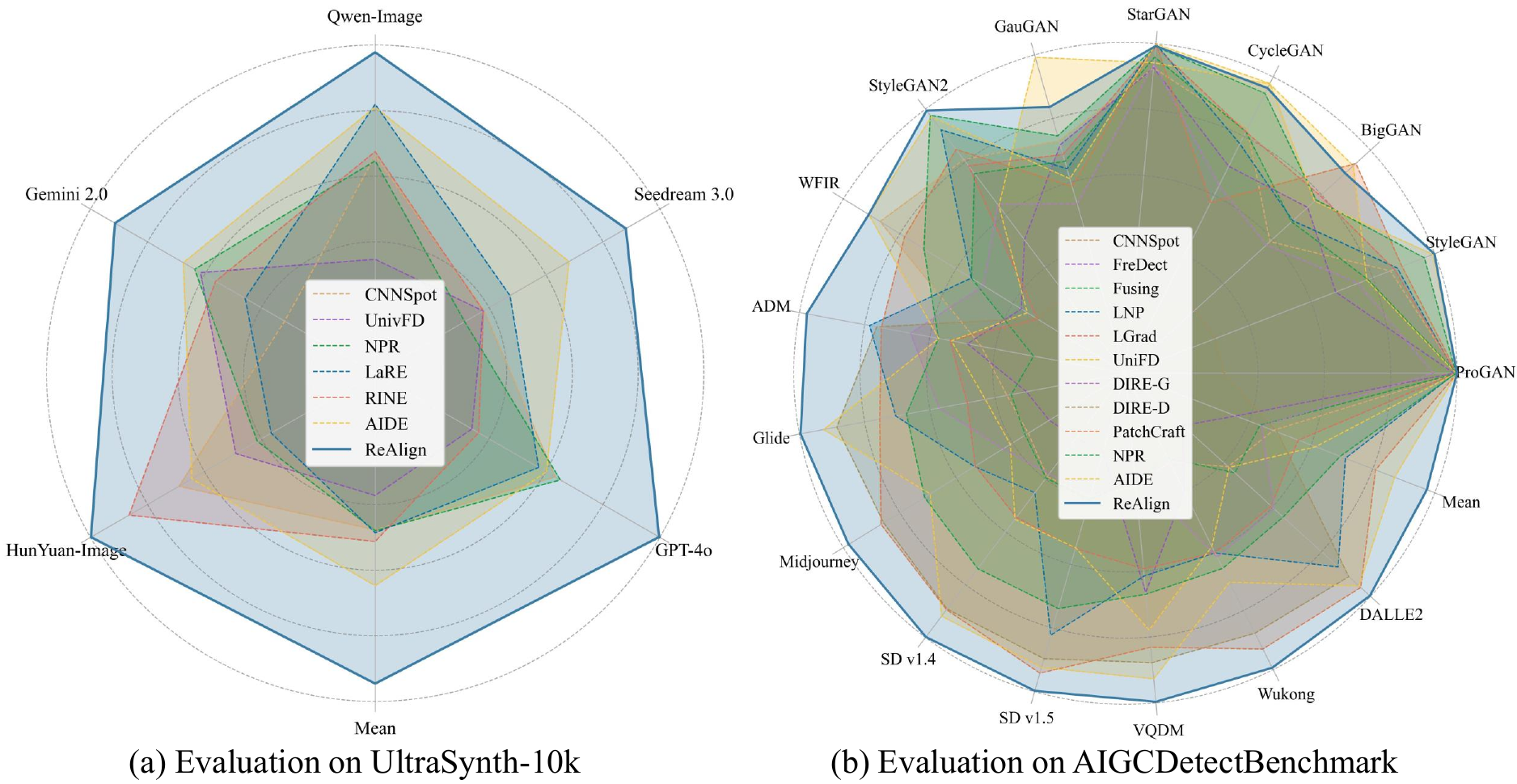}
	\vspace{-20pt}
	\caption{\textbf{Evaluation Result on UltraSynth-10k and AIGCDetectBenchmark~\cite{zhong2023patchcraft}.} Our ReAlign achieves SOTA performance.}
	\label{leida}
    \vspace{-10pt}
\end{figure}

Nowadays, state-of-the-art (SOTA) AIGI detection methods have achieved remarkable performance. These approaches can generally be categorized into two types based on whether they incorporate large language models (LLMs): non-LLM-based~\cite{li2025texture,yang2025all,yan2024generalizing,zhou2024freqblender,cao2022end} and LLM-based~\cite{xu2024fakeshield,kang2025legion,qu2024textsleuth,zhang2025badwindtunnel,yao2025depthssc,yao2024swift,xiao2024confusion} methods.
Non-LLM-based approaches typically utilize CNNs or Vision Transformers (ViTs) to extract image features and produce binary classification results. However, such methods often operate as black boxes and tend to overfit the training data due to their limited parameter capacity.
With the advancement of LLMs~\cite{liu2024visual, Qwen2.5-VL,openai2023gpt}, researchers have begun to introduce them into synthetic image detection to overcome the limitations of traditional methods, giving rise to a series of LLM-based detection frameworks~\cite{liu2024forgerygpt,xu2024fakeshield}. These models can provide textual explanations for their decisions, offering greater transparency and interpretability throughout the detection process. Moreover, by inheriting the world knowledge embedded in LLMs, they exhibit enhanced sensitivity to semantically relevant forgery.
Notably, with the growing development of reinforcement learning (RL), especially optimization techniques for reasoning-based LLMs (e.g., GRPO, Group Relative Policy Optimization~\cite{guo2025deepseek}), an increasing number of researchers~\cite{xu2025avatarshield} are exploring their integration into synthetic detection tasks as an alternative to conventional supervised fine-tuning (SFT). Studies~\cite{chu2025sft,zhao2025reasoning} have shown that such outcome-driven reinforcement learning strategies generally achieve stronger generalization compared to SFT.
Although incorporating LLMs into synthetic detection helps overcome the black-box limitations of non-LLM-based methods and enables cross-domain detection, their drawbacks remain significant.
\textbf{First}, we observe that LLM-based detection methods are generally more adept at handling semantically errors related forgery, but less effective in dealing with low-level artifact-based forgeries.
For example, LLMs can easily identify regions that are factually inconsistent or semantically implausible within an image, yet they often struggle to detect simple low-level forgery traces like texture artifacts.
\textbf{Additionally}, these LLM-based methods are computationally intensive, with large parameter sizes, slow inference speed, and high deployment costs, making them unsuitable for deployment or inference on mobile or resource-constrained devices.
\textbf{Moreover}, although LLMs can generate explanatory text outputs, current studies have yet to provide clear evidence that such explanations contribute directly to improved detection performance.
Therefore, in this work, we investigate the intrinsic value of explanatory texts generated by LLMs on AIGI detection performance, and further explore the question: \textit{Is it possible to combine the strengths of both LLM-based and non-LLM-based AIGI detection methods, aiming to retain their advantages while mitigating their weaknesses?}

Through our experiments, we found that LLM-based detection methods are capable of producing a high-quality textual representation space, particularly those optimized with reinforcement learning. We further demonstrate that the reasoning textual representations possess the following three properties:
\textbf{(1) Discriminative Capability:} Textual representations generated by RL-optimized LLMs exhibit strong semantic correlation with the concepts of ``real'' and ``fake'', enabling effective forgery discrimination;
\textbf{(2) Cross-Domain Generalization:} Textual representations help bridge distribution gaps across datasets, showing robust domain invariance and improved generalization;
\textbf{(3) Semantic-Error Sensitivity:} Textual representations are sensitive to semantic inconsistencies while insensitive to underlying artifact details.

We leverage the unique properties of reasoning text representations to design ReAlign, an efficient and lightweight framework for AIGI forgery detection, which bridges the strengths of both LLM-based and non-LLM-based detection paradigms. By training AIGI-R1, a GRPO-optimized LLM-based detector, we extract highly discriminative and semantically rich reasoning representations. ReAlign leverages reasoning textual representations aligned with visual features through contrastive learning and a designed joint optimization strategy, achieving strong detection performance. 
Ultimately, ReAlign not only exhibits strong semantic-error sensitivity and cross-domain invariance, but also eliminates the need for large-scale models during inference, making it both lightweight and efficient.
Extensive experiments conducted on multiple benchmarks, including the newly constructed UltraSynth-10k, demonstrate that ReAlign exhibits SOTA detection performance and great generalization capability across diverse and complex generative models. Some of the experimental results are shown in Fig.~\ref{leida}.
Our contributions are summarized as follows:

\noindent \ding{113}~(1) We analyze the strengths and limitations of non-LLM-based and LLM-based AIGI detection methods, and identify that reasoning text serves as a discriminative, cross-domain, and semantic-error sensitive representation.



\noindent \ding{113}~(2) We propose AIGI-R1, a reasoning-based multimodal large model for AIGI detection built upon GRPO, and introduce a novel framework for constructing reasoning text representation, which can be further utilized for alignment.


\noindent \ding{113}~(3) We propose ReAlign, a parameter-efficient training framework that aligns the text space of a vision-language foundation model with a forgery-aware reasoning text space. This enables training a lightweight and highly generalizable forgery detection model.

\noindent \ding{113}~(4) We construct a new dataset UltraSynth-10k, including 5 SOTA image generation methods. ReAlign shows excellent detection capabilities and stable generalization performance across diverse AIGI benchmarks.



\section{Related Works}
\label{sec:formatting}

\begin{figure*}[t]
	\centering
	\includegraphics[width=0.95\linewidth]{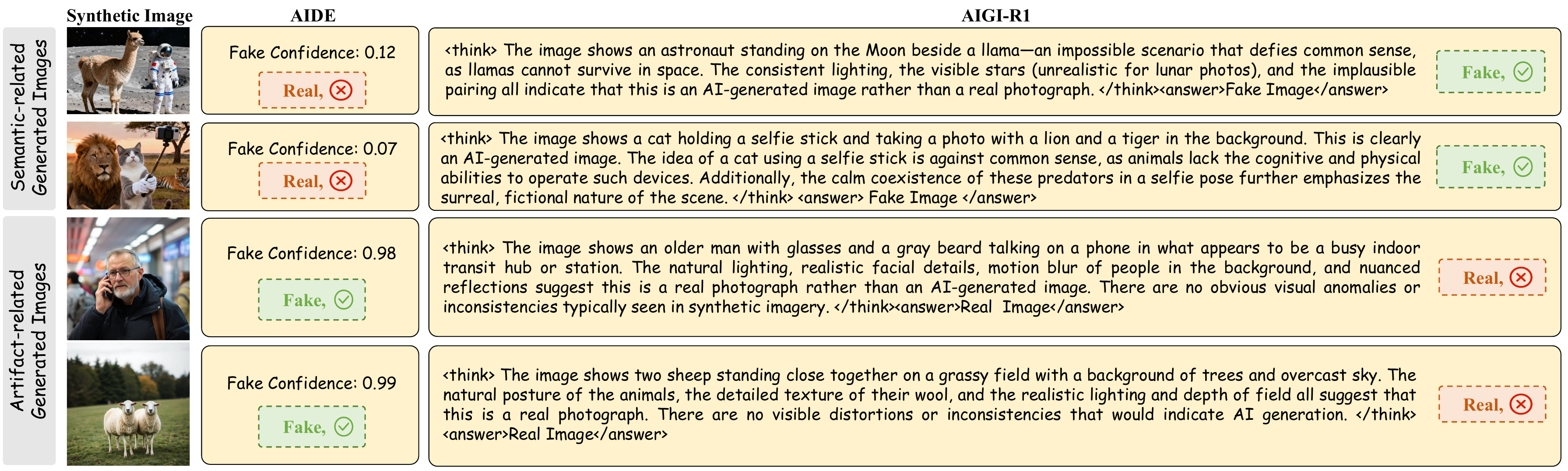}
	\vspace{-10pt}
	\caption{\textbf{A study comparing LLM-based detectors and non-LLM-based detectors on different types of forgeries.} We select AIDE as the non-LLM-based detection method, and AIGI-R1 as the LLM-based detection method.}
	\label{yuyi}
    \vspace{-15pt}
\end{figure*}
%
\subsection{Non-LLM-based AIGI Detection}

With the rapid advancement of AIGC technologies~\cite{gao2025seedream,comanici2025gemini,zhang2025exploit}, particularly diffusion models and deepfake techniques, the forms of image forgery have become increasingly diverse, extending from early facial forgeries to various domains and applications. This growing complexity not only disrupts the real-world information environment but also creates an urgent need for effective forgery detection.

In recent years, researchers have proposed a variety of methods~\cite{ma2023iml,zhong2023rich,salvi2023robust,zhou2025aigi,lin2025seeing,huang2025unishield,du2025forensichub,zhang2024editguard,zhang2025omniguard} to extract low-level forgery cues, typically focusing on RGB enhancement, frequency information, reconstruction residuals, gradient features, and inter-pixel relationships. Representative works include LGrad~\cite{tan2023learning} used gradient maps generated by a classifier as features for GAN detection.
UniFD~\cite{ojha2023fakedetect} firstly utilized the vision-language model CLIP for high-level semantic extraction.
FatFormer~\cite{luo2024forgery} introduced a frequency-aware adapter into CLIP to boost sensitivity to frequency-domain artifacts. 
PatchCraft~\cite{zhong2023patchcraft} compared pixel correlations between texture-rich and poor regions to extract universal features.
OpenSDI~\cite{wang2025opensdi} integrated low-level frequency clues with CLIP’s semantic embeddings.
C2P-CLIP~\cite{tan2025c2p} enhanced discriminative understanding via contrastive learning on class-guided prompts. 
AIDE~\cite{yan2024sanity} proposed a hybrid model that jointly modeled low- and high-level features to capture semantic and frequency differences between real and generated images. Although these methods show strong performance in artifact detection, they often face limited generalization when encountering unseen generative models or domains. This is mainly due to their reliance on fixed visual features and insufficient semantic adaptability. Improving generalization across generators, domains, and modalities remains a key challenge.


\subsection{LLM-based AIGI Detection}

With the development of multimodal large language models (MLLMs)~\cite{dai2023instructblip, liu2024visual, chen2023sharegpt4v,zhang2025vq,lan2025contextual,zeng2025janusvln,li2025lion,xiaoreversible}, models have shown the ability to project image content into language space and generate explanatory texts, enabling natural multimodal interaction and visual reasoning. Building on this, researchers have explored MLLMs for image forgery detection and interpretation.

FakeShield~\cite{xu2024fakeshield} first introduced LLMs into forgery detection, proposing an explainable framework that supports fine-grained detection, localization, and interpretation.  
ForgeryGPT~\cite{liu2024forgerygpt} enhanced forgery detection by modeling high-order forensic knowledge across multilingual spaces and using a customized LLM for interactive reasoning.  
MM-IML~\cite{huang2025mm} presented a general-specialized multimodal framework that fused MLLM semantic understanding with expert models via cross-modal tampering features.  
SIDA~\cite{huang2024sida} focused on social media scenarios, introducing SID-Set and a dual-label mechanism to improve robustness in complex environments.  
LEGION~\cite{kang2025legion} proposed a detection–generation collaboration framework on the SynthScars dataset, enabling both localization and guided content restoration.  
Additionally, Sofake~\cite{huang2025so} and AvatarShield~\cite{xu2025avatarshield} leveraged reasoning-based MLLMs for forgery interpretation and consistency analysis, pushing forward cross-modal generalization. Despite their progress, most LLM-based approaches remain computationally expensive and struggle with low-level artifact detection, leaving room for improvement in both efficiency and sensitivity.

\section{Methodology}

\begin{figure*}[t]
	\centering
	\includegraphics[width=1.0\linewidth]{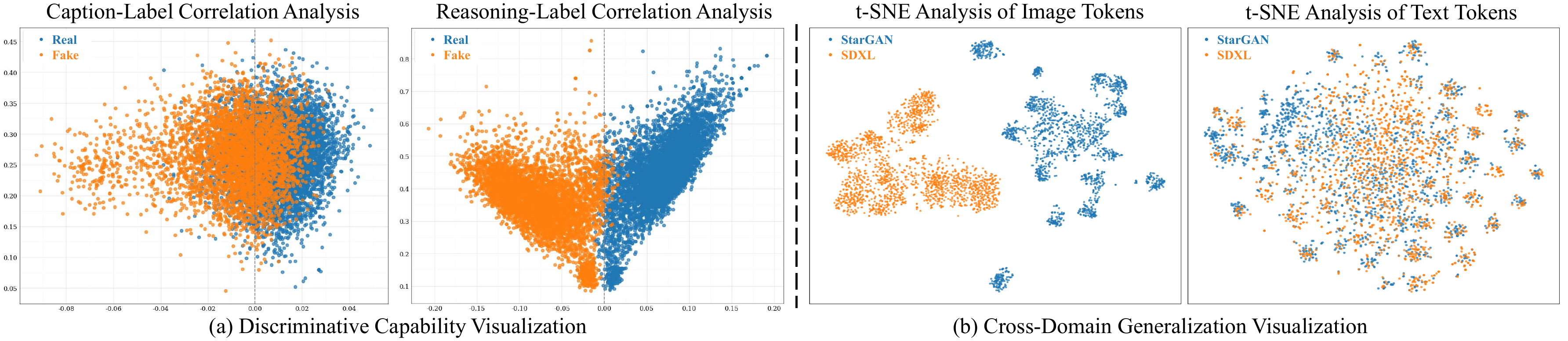}
	\vspace{-18pt}
	\caption{\textbf{Visualization of the discriminative capability and generalization properties of reasoning text representations.}}
	\label{tsne}
    \vspace{-10pt}
\end{figure*}

\subsection{Analysis and Motivation}
We first rethink the mechanisms behind non-LLM-based and LLM-based forgery detection approaches.

\noindent \textbf{Non-LLM-based:} The non-LLM-based methods~\cite{ma2023iml,yan2024sanity} mainly rely on CNNs or ViTs to extract image features, focusing on pixel-level anomalies such as noise irregularities, and frequency artifacts. Given an input image $\mathbf{x}_i$ and its real/fake label $y_i \in \{0,1\}$, the detector $f_{\theta}(\cdot)$ outputs a probability $\hat{p}_i = f_{\theta}(\mathbf{x}_i) \in (0,1)$. The model is optimized with a standard binary cross-entropy (BCE) loss:
\begin{equation}
    \mathcal{L}_{\text{BCE}}
    = - \frac{1}{N} \sum_{i=1}^{N} \big[ 
        y_i \log \hat{p}_i + (1 - y_i) \log (1 - \hat{p}_i)
    \big],
    \label{eq:bce}
\end{equation}
where $N$ is the batch size. The BCE loss provides a clear discriminative signal that explicitly pushes the encoder to separate real and fake images in the feature space. By penalizing prediction errors based on class labels, it guides the visual encoder to adjust embeddings along directions that maximize linear separability between the two classes. Furthermore, because the BCE objective focuses on classification confidence rather than semantic reconstruction, the strongest gradients naturally arise from local inconsistencies, such as boundary misalignment, texture discontinuity, or noise artifacts.


\noindent \textbf{LLM-based:} In contrast, LLM-based AIGI detectors~\cite{lin2025seeing,zhou2025aigi,xu2025avatarshield} encode images into visual tokens and fuse them with textual instructions before feeding them to the LLM. The model then generates both reasoning text within the $<$think$>$ and $<$/think$>$ tags and the judgement answer within the $<$answer$>$ and $<$/answer$>$. Taking SFT as an instance, the model learns through next-token prediction on instruction-response pairs, e.g., \textit{``Determine whether the image is real or AI-generated and explain your reasoning.''} The training objective is formulated as:
\begin{equation}
\mathcal{L}_{\text{LLM}} = - \sum_{t=1}^{T} \log P_{\Theta}(\mathbf{w}_t \mid \mathbf{w}_{<t}, \mathbf{V}, \mathbf{C}),
\end{equation}
where $\mathbf{w}_t$, $\mathbf{V}$ and $\mathbf{C}$ denote the predicted tokens, visual tokens and textual instruction. Such token-level supervision emphasizes contextual consistency between visual evidence and textual responses, but lacks explicit constraints on fine-grained visual separability. As a result, the LLM tends to interpret forgery parts as a semantic anomaly (e.g., an illogical object) rather than a low-level artifact (e.g., edge discontinuity or blending noise). 
Meanwhile, the large-scale instruction-tuning datasets~\cite{jia2021scaling,schuhmann2022laion} used for pretraining and alignment predominantly include high-level visual reasoning tasks, such as captioning, common sense judgment, while rarely exposing the model to pixel-level forgeries. This semantic bias in data distribution further drives the model to prioritize meaning over microscopic visual inconsistencies, limiting its sensitivity to subtle forgery traces.

Thus, both types of forgery detection methods have their own strengths.
Traditional non-LLM-based approaches typically rely on low-level visual feature extraction and are trained on specific types of forgery datasets, enabling them to achieve high accuracy in detecting artifact-based manipulations.
In contrast, LLM-based methods possess superior semantic understanding, allowing them to identify potential anomalies based on logical inconsistencies and common-sense reasoning within the image content, as shown in Fig.~\ref{yuyi}. 
This raises a key question: \textit{Can we integrate the strengths of LLM-based and non-LLM-based approaches into a unified framework?}


\begin{figure*}[t]
	\centering
	\includegraphics[width=0.95\linewidth]{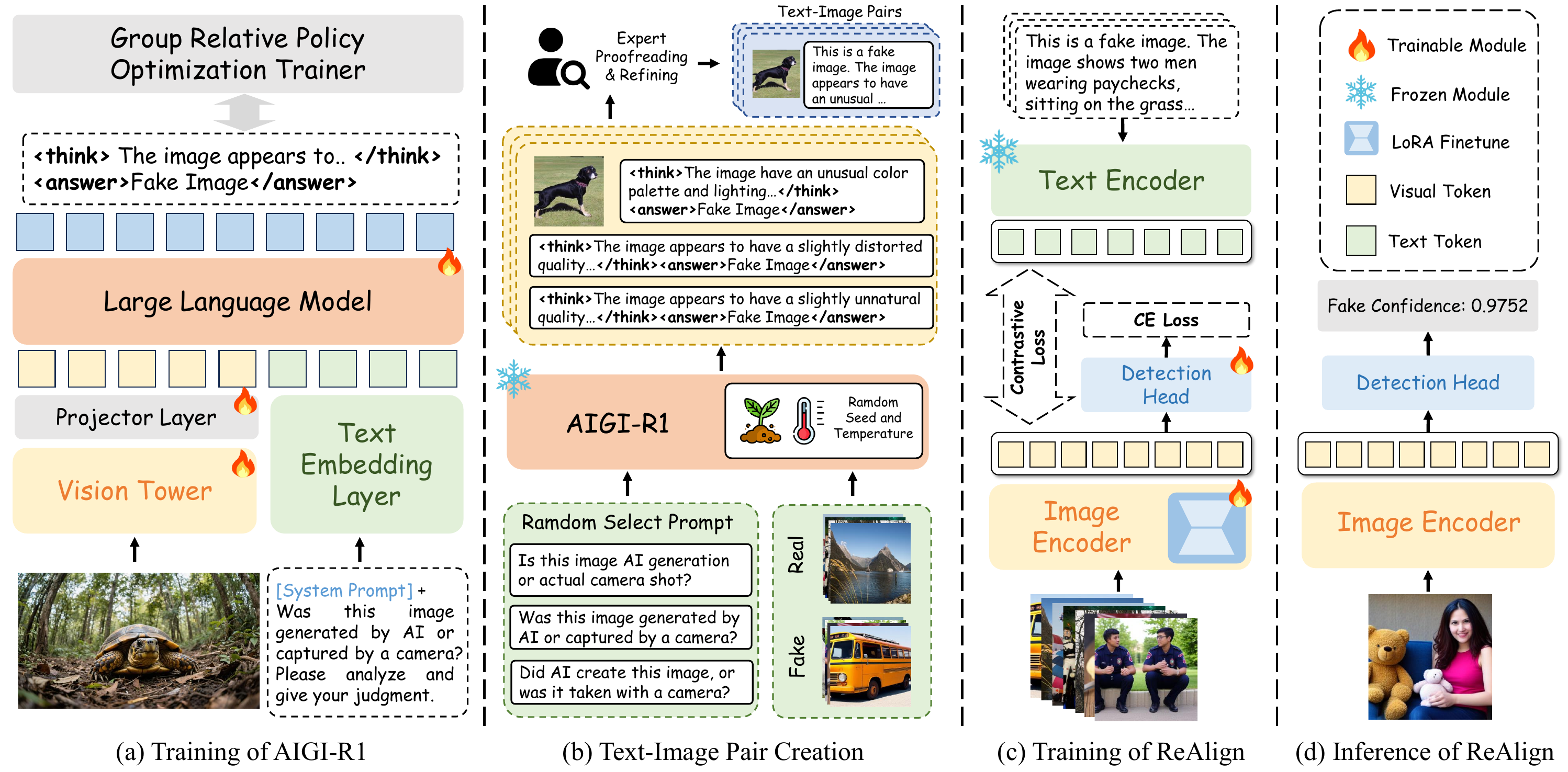}
	\vspace{-10pt}
	\caption{\textbf{The pipeline of ReAlign.} (a) The GRPO optimization pipeline of AIGI-R1. (b) Reasoning texts are collected from the trained AIGI-R1 and paired with the corresponding images to form a text-image pairs dataset. (c) Joint training of alignment and classification tasks for ReAlign based on the collected text-image dataset. (d) Using the trained ReAlign model for AIGI detection.}
	\label{pipeline}
    \vspace{-15pt}
\end{figure*}

\subsection{Training of AIGI-R1}



\label{grpo}
To better understand the characteristics of the LLM-based AIGI detector, and inspired by DeepSeek-R1~\cite{guo2025deepseek, shen2025vlm}, we propose optimizing the MLLM via outcome-based reinforcement learning in the R1 paradigm, and introduce AIGI-R1. Group Relative Policy Optimization (GRPO)~\cite{guo2025deepseek} extends Proximal Policy Optimization (PPO) as a more advanced form of reinforcement learning. Rather than estimating values through a critic as PPO does, GRPO collectively assesses several candidate responses by measuring their relative rewards, which removes the need for explicit value modeling. Through this mechanism, training becomes more stable and efficient, making GRPO especially advantageous for tasks with scarce supervision or those where the quality of outcomes is best judged through comparison. Furthermore, recent studies~\cite{huang2025so} have shown that, in forgery detection, GRPO can better stimulate the model’s generalization ability compared to SFT, ensuring more reliable performance on unseen data.
The optimization objective of GRPO is defined as:
\vspace{-17pt}

\begin{align}
\label{GRPO}
&\max_{\pi_\theta} \;  \mathbb{E}_{o \sim \pi_\theta(q)} [ R_{\text{GRPO}}(q, o) ] \\
&=[ R_{\text{total}}(q, o) - \beta\cdot\mathrm{KL}[ \pi_\theta(o|q)~\| ~\pi_{\text{ref}}(o|q)]],
\end{align}
where $\pi_{\text{ref}}$ is the reference model prior to optimization, $R_{\text{GRPO}}$ is the relative reward function used to compare candidates, $\beta$ is the hyperparameter controlling the KL divergence, and $R_{\text{total}}(q, o)$ is the relative reward function that evaluates candidate outputs. The reward is defined as:
\vspace{-5pt}

\begin{equation}
R_{\text{total}}(q, o) = R_{\text{det}}(q, o)+R_{\text{format}}(q,o), \end{equation}
\begin{equation}
R^{(i)}_{\text{det}}(q,o) = 1, \quad \text{if } o^{(i)} = \text{det}_{\text{gt}}~\text{ else } 0,
\end{equation}
where $R_{\text{det}}$ is the detection accuracy reward, and $R_{\text{format}}$ is the formatting reward. 
During the optimization process, we directly use the image class labels (``real" or ``fake") as the ground truth for the answers and formulate the question as: \textit{``Was this image generated by AI or captured by a camera? Please analyze and give your judgment."} 
We also design a detailed system prompt during the training process to guide the MLLM in observing details and to inspire it to perform AIGI detection.
We feed the image, system prompt, and question into the MLLM together. The model architecture and training process are illustrated in Fig.~\ref{pipeline}(a). More details are provided in the supplementary material.


\subsection{Reasoning Text as a Bridge}
\label{bridge}

We argue that the core reason LLM-based methods possess forgery detection capabilities stems from the representation space constituted by the reasoning text. By utilizing the reasoning text space, we can bridge the gap between LLM-based and non-LLM-based detection methods. Through a series of experiments, we validate three key properties of this reasoning representation space.


\noindent \textbf{Discriminative Capability:} To investigate the discriminative capability of the reasoning text representation, we select a subset of real and AI-generated images~\cite{zhong2023patchcraft} and compare their image captions produced by Qwen2.5-VL~\cite{Qwen2.5-VL} with reasoning texts generated by AIGI-R1. We compute the semantic similarity of these two different texts to two class labels, ``real'' and ``fake'', denoted as $s_\text{real}$ and $s_\text{fake}$, respectively. Each sample is then projected onto a 2D plane, with the $x$-axis representing the directional bias toward one class ($s_\text{real} - s_\text{fake}$), and the $y$-axis representing the relevance to the real or fake concept ($s_\text{real} + s_\text{fake}$).
As shown in Fig.~\ref{tsne}(a), the reasoning texts exhibit a stronger polarity between the real and fake clusters along the $x$-axis, whereas the caption texts show a larger degree of overlap. This indicates that the reasoning texts provide a stronger discriminative signal. Moreover, the reasoning texts have a higher mean vertical distribution along the $y$-axis, suggesting that their semantic representations are more closely aligned with the real/fake discrimination space.


\noindent \textbf{Cross-Domain Generalization:} To investigate the role of reasoning text representations in mitigating cross-domain differences, we visualize both the visual features and the textual features generated by AIGI-R1 from the StarGAN and SDXL datasets~\cite{zhong2023patchcraft} using t-SNE~\cite{maaten2008visualizing}, as shown in Fig.~\ref{tsne}(b). The left plot shows the distribution of images in the visual representation space. It is evident that the two datasets form almost entirely separate clusters, indicating a substantial distributional shift in the visual modality.
In contrast, the right plot presents the distribution of reasoning texts generated by the reasoning-based LLM in the textual representation space. Here, we observe a high degree of overlap between the two datasets, suggesting that reasoning texts exhibit greater domain invariance in their textual representations. This indicates that textual features can effectively bridge domain gaps, thereby enhancing model robustness and generalization in cross-dataset scenarios.


\noindent \textbf{Semantic-Error Sensitivity:} As shown in Fig.~\ref{yuyi}, we compare a non-LLM-based method AIDE~\cite{yan2024sanity} with our reasoning-based AIGI-R1. The results show that AIDE performs well in detecting artifact-related forgeries, where visual inconsistencies, such as texture distortions, are present. However, it fails to identify semantically related forgeries that violate logical or common-sense reasoning. In contrast, AIGI-R1 excels at detecting such semantic anomalies but struggles with subtle low-level visual artifacts.

\subsection{ReAlign}


Based on the above analysis, we have a clear understanding of the advantages of reasoning-based LLM detectors and the value of reasoning text representation. In this section, we further explore how to effectively transfer the detection capability of LLMs to lightweight small models such as CLIP. To achieve this demand, we propose a method called ReAlign, whose core idea focuses on constructing alignment training pairs and designing an alignment framework.

\noindent \textbf{Constructing Text-Image Training Pairs.} Specifically, we employ the pre-trained AIGI-R1 to construct text-image pairs, as illustrated in Fig.~\ref{pipeline}(b). To enhance the diversity of textual data, we input multiple questions for each image and generate multiple corresponding answers. Furthermore, by adjusting the seed and temperature during the prediction process, we further increase response variability. After obtaining the generated answers, human experts are invited to verify and refine the outputs. Then, the reasoning text enclosed within the $<$think$>$ and $<$/think$>$ tags is extracted, and a prefix ``This is a real/fake image.'' is added according to the image label. 
Finally, the refined forgery description texts are paired with their corresponding images, creating a high-quality text-image pairs dataset with strong forgery semantic features. 

\noindent \textbf{Reasoning Alignment and Inference Process.} Building on the constructed text-image pairs, as illustrated in Fig.~\ref{pipeline}(c), ReAlign leverages contrastive learning to finetune a pretrained CLIP model, enhancing its generalization ability and sensitivity to semantic errors. The ReAlign detection model is composed of three modules: an image encoder, a text encoder, and a detection head. We initialize both the image and text encoders with pretrained CLIP~\cite{radford2021learning}, keeping the text encoder frozen while efficiently finetuning the image encoder using the LoRA strategy. This allows the CLIP model to align with the reasoning representations while preserving its inherent general semantic understanding capabilities. Following the image encoder, a two-layer MLP subnet is added to perform the binary classification task for detecting real or fake images.

To ensure that ReAlign simultaneously aligns the reasoning text space and maintains strong downstream detection performance, we jointly optimize the CLIP using both a contrastive loss and a detection loss. Specifically, the contrastive loss is a symmetric cross-entropy loss composed of $\mathcal{L}_{i \rightarrow t}$ and $\mathcal{L}_{t \rightarrow i}$. Taking $\mathcal{L}_{i \rightarrow t}$ as an example:
\begin{equation}
    \label{i2t}
    \mathcal{L}_{i \rightarrow t} = -\frac{1}{N} \sum_{i=1}^{N} \log \frac{\exp(\mathbf{v}_i \cdot \mathbf{t}_i)}{\sum_{j=1}^{N} \exp(\mathbf{v}_i \cdot \mathbf{t}_j)},
\end{equation}
where $\mathbf{v}_i$ and $\mathbf{t}_i$ represent the vectors obtained by encoding a forged image and its corresponding forged text embeddings produced by the text encoder, respectively. Similarly, $\mathcal{L}_{t \rightarrow i}$ can be derived in the same way. Finally, the contrastive loss can be expressed as follows.
\begin{equation}
    \label{eq:alignloss} 
    \mathcal{L}_{\text{contrastive}} = \frac{1}{2} \left( \mathcal{L}_{i \rightarrow t} + \mathcal{L}_{t \rightarrow i} \right).
\end{equation}

It works by pulling together semantically consistent image-text pairs while pushing apart mismatched pairs, thus building a consistent cross-modal embedding space. The detection loss is computed using the binary cross-entropy between the predicted image classification results and the ground-truth labels, as shown in Eq.~\ref{eq:bce}. The final loss function is obtained by the weighted sum of the above loss functions, formulated as:
\begin{equation}
    \label{t2i} \quad 
    \mathcal{L} = \mathcal{L}_{\text{contrastive}} + \alpha \cdot \mathcal{L}_{\text{classification}},
\end{equation}
where $\mathcal{\alpha}$ is the hyperparameter and set to 8. \textbf{During inference}, as shown in Fig.~\ref{pipeline}~(d), only the image encoder and the detection head are required to produce accurate predictions, without the need to explicitly include the reasoning process.

\section{Experiment}

\begin{table*}[t]
\vspace{-3mm}
\caption{\textbf{Performance Comparison on the AIGCDetectBenchmark.} The best result and the second-best result are marked in \textbf{bold} and \underline{underline}, respectively. The rows represent different synthetic image generation methods, and the columns represent comparison detection methods. The metrics in the table are accuracy(\%).}
\begin{center}
\vspace{-20pt}
\fontsize{13}{13}\selectfont 
\resizebox{1.0\linewidth}{!}{  
\begin{tabular}{lccccccccccc|cc}
\toprule
\textbf{Method} & \textbf{CNNSpot\cite{wang2019cnnspot}} & \textbf{FreDect\cite{frank2020fredect}} & \textbf{Fusing\cite{ju2022fusing}} & \textbf{LNP\cite{liu2022detecting}} & \textbf{LGrad\cite{tan2023learning}} & \textbf{UniFD\cite{ojha2023fakedetect}} & \textbf{DIRE-G\cite{wang2023dire}} & \textbf{DIRE-D\cite{wang2023dire}} & \textbf{PatchCraft\cite{zhong2023patchcraft}} & \textbf{NPR\cite{tan2024rethinking}} & \textbf{AIDE\cite{yan2024sanity}} & \textbf{AIGI-R1} & \textit{\textbf{ReAlign}} \\
\midrule
ProGAN & \textbf{100.00} & 99.36 & \textbf{100.00} & 99.67 & 99.83 & 99.81 & 95.19 & 52.75 & \textbf{100.00} & 99.79 & \underline{99.99} & 95.83 & \textbf{100.00} \\
StyleGAN & 90.17 & 78.02 & 85.20 & 91.75 & 91.08 & 84.93 & 83.03 & 51.31 & 92.77 & 97.70 & \textbf{99.64} & 78.84 & \underline{97.93}\\
BigGAN & 71.17 & 81.97 & 77.40 & 77.75 & 85.62 & \underline{95.08} & 70.12 & 49.70 & \textbf{95.80} & 84.35 & 83.95 & 80.50 & 90.73 \\
CycleGAN & 87.62 & 78.77 & 87.00 & 84.10 & 86.94 & \underline{98.33} & 74.19 & 49.58 & 70.17 & 96.10 & \textbf{98.48} & 89.63 & 95.35 \\
StarGAN & 94.60 & 94.62 & 97.00 & \underline{99.92} & 99.27 & 95.75 & 95.47 & 46.72 & \textbf{99.97} & 99.35 & 99.91 & 78.31 & 97.47 \\
GauGAN & 81.42 & 80.57 & 77.00 & 75.39 & 78.46 & \textbf{99.47} & 67.79 & 51.23 & 71.58 & 82.50 & 73.25 & 82.07 & \underline{86.71} \\
StyleGAN2 & 86.91 & 66.19 & 83.30 & 94.64 & 85.32 & 74.96 & 75.31 & 51.72 & 89.55 & \textbf{98.38} & \underline{98.00} & 82.96 & 97.62 \\
WFIR & \underline{91.65} & 50.75 & 66.80 & 70.85 & 55.70 & 86.90 & 58.05 & 53.30 & 85.80 & 65.80 & \textbf{94.20} & 57.90 & 84.45 \\
ADM & 60.39 & 63.42 & 49.00 & 84.73 & 67.15 & 66.87 & 75.78 & \textbf{98.25} & 82.17 & 69.69 & 93.43 & \underline{97.89} & 97.80 \\
Glide & 58.07 & 54.13 & 57.20 & 80.52 & 66.11 & 62.46 & 71.75 & 92.42& 83.79 & 78.36 & 95.09 & \underline{96.85} & \textbf{97.76} \\
Midjourney & 51.39 & 45.87 & 52.20 & 65.55 & 65.35 & 56.13 & 58.01 & 89.45 & 90.12 & 77.85 & 77.20 & \textbf{96.87} & \underline{96.63} \\
SD v1.4 & 50.57 & 38.79 & 51.00 & 85.55 & 63.02 & 63.66 & 49.74 & 91.24 & 95.38 & 78.63 & 93.00 & \textbf{98.56} & \underline{97.85} \\
SD v1.5 & 50.53 & 39.21 & 51.40 & 85.67 & 63.67 & 63.49 & 49.83 & 91.63 & 95.30 & 78.89 & 92.85 & \textbf{98.42} & \underline{97.79} \\
VQDM & 56.46 & 77.80 & 55.10 & 74.46 & 72.99 & 85.31 & 53.68 & 91.90 & 88.91 & 78.13 & 95.16 & \underline{95.22} & \textbf{97.78} \\
Wukong & 51.03 & 40.30 & 51.70 & 82.06 & 59.55 & 70.93 & 54.46 & 90.90 & 91.07 & 76.11 & 93.55 & \underline{97.48} & \textbf{97.83} \\
DALLE2 & 50.45 & 34.70 & 52.80 & 88.75 & 65.45 & 50.75 & 66.48 & 92.45 & \underline{96.60} & 64.90 & \underline{96.60} & 90.05 & \textbf{97.80} \\
\textit{Mean} & 70.78 & 64.03 & 68.38 & 83.84 & 75.34 & 78.43 & 68.68 & 71.53 & 89.31 & 82.91 & \underline{92.77} & 91.77 & \textbf{96.14} \\
\bottomrule
\end{tabular}
}
\end{center}
\label{table:b1}
\vspace{-3mm}
\end{table*}

\begin{table*}[t]
\vspace{-3mm}
\caption{\textbf{Performance Comparison on the AIGI-Holmes.} Our method can achieve the best detection accuracy.}
\begin{center}
\vspace{-20pt}
\fontsize{13}{13}\selectfont 
\resizebox{1.0\linewidth}{!}{  
\begin{tabular}{lcccccccc|cc}
\toprule
\textbf{Method} & \textbf{CNNSpot\cite{wang2019cnnspot}} & \textbf{AntiFakePrompt\cite{chang2023antifakeprompt}} & \textbf{UniFD\cite{ojha2023fakedetect}} & \textbf{NPR\cite{tan2024rethinking}} & \textbf{LaRE\cite{luo2024lare}} & \textbf{RINE\cite{koutlis2024leveraging}} & \textbf{AIDE\cite{yan2024sanity}} & \textbf{AIGI-Holmes\cite{zhou2025aigi}} & \textbf{AIGI-R1} & \textit{\textbf{ReAlign}} \\
\midrule
Janus & 70.00 & 72.20 & 87.60 & 51.20 & 70.80 & 89.90 & \underline{91.20} &  80.20 & 75.97 & \textbf{95.84} \\
J-Pro-1B & 70.90 & 84.30 & 96.90 & 69.50 & 74.70 & 98.70 & \underline{98.90} & 91.90 & 92.61 & \textbf{99.64} \\
J-Pro-7B & 85.00 & 84.80 & 96.40 & 73.90 & 95.60 & 97.20 & \underline{97.80} & 89.60 & 88.58 & \textbf{99.85} \\
Show-o & 72.20 & 86.20 & 85.90 & 93.70 & 80.00 & 98.80 & 98.00 & 98.00 & \underline{99.08} & \textbf{99.89} \\
LlamaGen & 61.90 & 96.20 & 93.10 & 93.50 & 91.60 & 99.10 & 99.40 & 98.00 & \underline{99.87} & \textbf{99.92} \\
Infinity & 86.80 & 83.60 & 79.20 & 93.80 & 77.90 & 99.20 & 98.70 & 98.40 & \underline{99.87} & \textbf{99.95} \\
VAR & 59.90 & 90.70 & 64.30 & 85.90 & \underline{98.80} & 85.00 & 93.60 & 76.00 & 80.36 & \textbf{99.74} \\
PixArt-XL & 78.20 & 81.70 & 75.70 & 93.40 & 82.20 & 98.90 & 98.60 & 98.50 & \underline{99.84} & \textbf{99.79} \\
SD3.5-L & 63.80 & 92.80 & 87.80 & 91.60 & 94.10 & 97.80 & 99.40 & 97.80 & \underline{99.68} & \textbf{99.83} \\
FLUX & 79.90 & 66.10 & 69.60 & 93.60 & 84.30 & \underline{97.10} & 94.40 & 94.20 & \underline{99.51} & \textbf{99.74} \\
\textit{Mean} & 72.90 & 83.90 & 83.60 & 84.00 & 85.00 & 96.20 & \underline{97.00} & 92.30 & 93.54 & \textbf{99.44} \\
\bottomrule
\end{tabular}
}
\end{center}
\label{table:b2}
\vspace{-20pt}
\end{table*}

\subsection{Experiment Setup}

\textbf{Implementation Details.}
We train the proposed AIGI-R1 using 8 NVIDIA A800 80GB GPUs. The learning rate is set to $1$$\times$$10^{-6}$ and $\beta$ coefficient is set to 0.04. The GRPO optimization process follows the R1-V~\cite{chen2025r1v} training framework to enhance the model’s reasoning and descriptive generation quality.
For the training of our ReAlign, we initialize the image encoder and text encoder via CLIP-ViT-L/14-336~\cite{radford2021learning}.
We freeze the text encoder and fine-tune the image encoder using LoRA (rank=$6$, alpha=$6$), while simultaneously training the detection head with full parameters. The model is trained for 10 epochs on a single NVIDIA RTX 3090 GPU, using a learning rate of $1$$\times$$10^{-4}$.


\noindent  \textbf{Datasets.}
To demonstrate the AIGI detection performance of our ReAlign method, we conduct a generalization evaluation on two mainstream benchmarks: AIGCDetectBenchmark\cite{wang2020cnn} and the AIGI-Holmes dataset\cite{zhou2025aigi}, following the same training settings and official evaluation protocols. In addition, we construct a challenging AIGI detection benchmark, UltraSynth-10k, using several advanced image generation methods. We also conduct a comprehensive comparison between ReAlign and other methods on this dataset. We evaluate the performance of our method using mean accuracy (mAcc) as a metric.

\noindent  \textbf{Comparison Methods.}
For all comparison experiments, we include AIGI-Holmes~\cite{zhou2025aigi} and our AIGI‑R1 as the LLM-based methods, while all remaining detectors~\cite{wang2019cnnspot,frank2020fredect,ju2022fusing,liu2022detecting,tan2023learning,ojha2023fakedetect,wang2023dire,zhong2023patchcraft,tan2024rethinking,yan2024sanity,chang2023antifakeprompt,luo2024lare,koutlis2024leveraging,koutlis2024leveraging} are categorized as non‑LLM‑based detection models, such as AIDE~\cite{yan2024sanity}, UniFD~\cite{ojha2023fakedetect}, NPR~\cite{tan2024rethinking}, and PatchCraft~\cite{zhong2023patchcraft}.

\subsection{Comparison to State-of-the-art Models}
\noindent \textbf{Evaluation on AIGCDetectBenchmark~\cite{zhong2023patchcraft}.} 
For fairness testing, all detection methods are fine-tuned on the training set of AIGCDetectBenchmark~\cite{zhong2023patchcraft} and evaluated on its test set, which contains diverse and challenging samples from 18 advanced generative models, as reported on Tab.~\ref{table:b1}.
ReAlign consistently achieves top performance across all models, with an average accuracy of 96.14\%, significantly outperforming prior methods such as AIDE (92.77\%) and our own GRPO-trained LLM baseline, AIGI-R1 (91.77\%), with gains of 3.37\% and 4.37\%, respectively.
Even on difficult high-fidelity models like SD v1.5, DALLE2, and Wukong, ReAlign maintains strong robustness with accuracies above 97\%. 
These results validate the effectiveness of GRPO-optimized reasoning texts and demonstrate the superior cross-domain capability and practicality of the ReAlign framework in AIGI detection.

\noindent \textbf{Evaluation on AIGI-Holmes Dataset~\cite{zhou2025aigi}.} 
Compared with AIGCDetectBenchmark, the AIGI-Holmes dataset is more challenging because it includes more advanced autoregressive-based and diffusion-based image generation methods, such as Infinity and FLUX. All the detectors are trained and tested on AIGI-Holmes Dataset~\cite{zhou2025aigi}, as shown in Tab.~\ref{table:b2}. Our ReAlign method continues to demonstrate a clear advantage, achieving the best detection performance across all compared detectors on the AIGI-Holmes benchmark. This dataset includes diverse and highly capable recent generative models, making it a strong test of generalization and practical robustness.

Compared to the second strong baseline AIDE (97.00\%), ReAlign improves the average performance by 2.44 percentage points. Even on challenging samples from advanced models such as SD3.5-L and FLUX, ReAlign achieves high accuracies of 99.83\% and 99.74\%, consistently outperforming SOTA methods like AIDE, RINE, and LaRE, and demonstrating strong robustness.
These results show that ReAlign delivers unparalleled advantages even against more challenging, novel generative models, further validating the effectiveness of our approach.

\subsection{UltraSynth-10k}

\begin{figure}[t]
	\centering
	\includegraphics[width=1.0\linewidth]{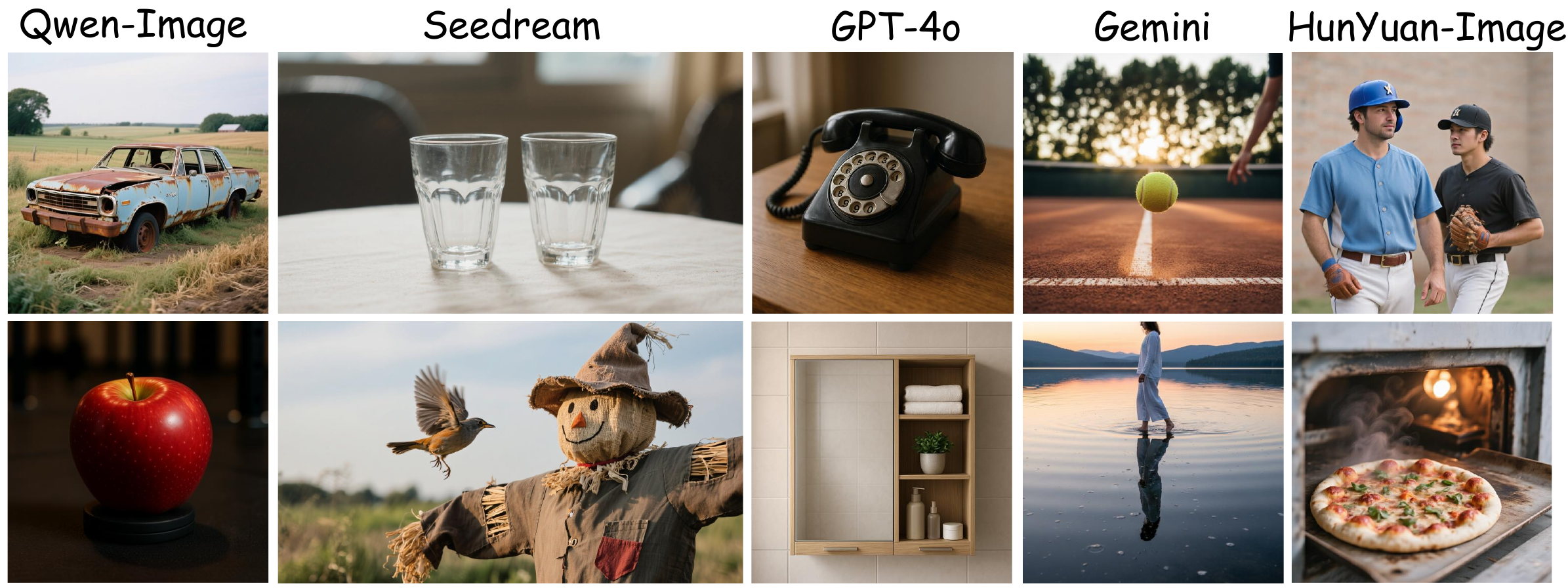}
	\vspace{-20pt}
	\caption{\textbf{Sampled Examples of UltraSynth-10k.}}
	\label{UltraSynth}
    \vspace{-10pt}
\end{figure}

Currently, the field of image generation is developing rapidly, with many advanced proprietary generation methods~\cite{gao2025seedream,comanici2025gemini,HunyuanImage-2.1,wu2025qwen} emerging. As shown in Fig.~\ref{UltraSynth}, the fake images generated by these methods have significantly surpassed many traditional open-source approaches in terms of realism and detail fidelity. However, existing AIGI detection benchmarks~\cite{zhong2023patchcraft,zhou2025aigi} still primarily focus on earlier or more common generation models, lacking comprehensive coverage of the latest and most powerful proprietary generation techniques. To fill this gap, we have collected a large number of high-quality fake images generated by advanced proprietary algorithms~\cite{gao2025seedream,comanici2025gemini,HunyuanImage-2.1,wu2025qwen,chen2023sharegpt4v} and constructed a challenging AIGI detection benchmark, UltraSynth-10k, which contains 10,000 real and fake images. 


We reused the weights of ReAlign and several comparison methods which are trained on AIGI-Holmes Dataset \textbf{without exposing to any generation methods} on UltraSynth-10k. The testing results of UltraSyth-10k are reported on Tab.~\ref{table:b3}. Our method ReAlign outperforms existing approaches on most sub-datasets, achieving the highest overall accuracy of \textbf{97.09\%}.
Even on datasets where other models perform well (e.g., RINE, where HunYuan-Image scores 92.56\%), ReAlign maintains a leading accuracy of 99.91\%. These results demonstrate ReAlign’s strong generalization ability and practical value in real-world applications. 
Notably, our AIGI-R1 also achieves an average \textbf{96.42\%} detection accuracy across the five SOTA generation methods, and even surpasses ReAlign on Seeddream and Gemini. This strongly shows the superior generalization ability of reasoning-based LLMs for detecting unseen generative models, and further validates the source of ReAlign’s own generalization capability.


\begin{table}[t]
\caption{\textbf{Performance Comparison on the UltraSynth-10k.}}
\begin{center}
\vspace{-20pt}
\fontsize{13}{13}\selectfont
\resizebox{1.0\linewidth}{!}{
\begin{tabular}{lcccccccc}
\toprule
\multirow{2}{*}{\textbf{Method}} & \textbf{Qwen-} & \textbf{Seedream} & \textbf{GPT-4o} & \textbf{Gemini} & \textbf{HunYuan-} & \multirow{2}{*}{\textbf{\textit{Mean}}} \\
& \textbf{Image~\cite{wu2025qwen}} & \textbf{\cite{gao2025seedream}} & \textbf{\cite{chen2023sharegpt4v}} & \textbf{\cite{comanici2025gemini}} & \textbf{Image~\cite{HunyuanImage-2.1}} \\
\midrule
CNNSpot & 80.54 & 60.68 & 78.36 & 57.07 & 82.98 & 71.93 \\
UniFD & 62.40 & 60.50 & 61.12 & 75.70 & 72.05 & 66.35 \\
NPR & 79.66 & 56.35 & 79.42 & 76.99 & 67.92 & 72.07 \\
LaRE & \underline{89.59} & 66.49 & 74.95 & 65.80 & 65.23 & 72.41 \\
RINE & 81.28 & 60.53 & 62.50 & 72.35 & 92.56 & 73.84 \\
AIDE & 89.10 & 79.72 & 76.56 & 79.50 & 80.50 & 81.08 \\
\midrule
AIGI-R1 & 89.27 & \textbf{95.33} & \underline{98.22} & \textbf{99.67} & \underline{99.59} & \underline{96.42} \\ 
\textit{\textbf{ReAlign}} & \textbf{98.73} & \underline{92.40} & \textbf{99.86} & \underline{94.58} & \textbf{99.91} & \textbf{97.09} \\
\bottomrule
\end{tabular}
}
\end{center}
\label{table:b3}
\vspace{-10pt}
\end{table}

\begin{table}[t]
\centering
\caption{\textbf{Accuracy (\%) under Different Text Input.} ``Class Label'' represents ``This is a real/fake image.'' ``Caption'' refers to the image caption generated by Qwen2.5-VL~\cite{Qwen2.5-VL}. ``Reasoning Text'' refers to the reasoning text generated by AIGI-R1.}
\begin{center}
\vspace{-20pt}
\resizebox{1\linewidth}{!}{ 
\begin{tabular}{c|ccc|c}
\toprule
\textbf{Case} & \textbf{Class Label} & \textbf{Image Caption} & \textbf{Reasoning Text} & \textbf{Mean Accuracy} \\
\hline
Ours & \checkmark &              & \checkmark   & 97.09 \\
\hline
(a) & \checkmark & \checkmark   &              & 91.63 \textcolor{DarkGreen}{(-5.46\%)} \\
(b) &            &              & \checkmark   & 96.87 \textcolor{DarkGreen}{(-0.22\%)} \\
(c) &            & \checkmark   &              & 88.32 \textcolor{DarkGreen}{(-8.77\%)} \\
(d) & \checkmark &              &              & 91.33 \textcolor{DarkGreen}{(-5.76\%)} \\
\bottomrule
\end{tabular}
}
\end{center}
\label{table:ablation}
\vspace{-20pt}
\end{table}

\subsection{Ablation Study}



\noindent  \textbf{Ablation Study on Alignment Text.} To assess the effect of the reasoning texts in ReAlign, we perform an ablation study on UltraSynth-10k by varying the alignment text. As shown in Tab.~\ref{table:ablation}, our full setting combines reasoning texts from AIGI-R1 with a simple prefix including class label, achieving the highest accuracy of 97.09\%. 
In case (a), replacing reasoning texts with image captions produced by Qwen2.5-VL leads to a notable drop of 5.46\%, which proves the value of our reasoning text.
Case (b), which uses reasoning texts alone, achieves 96.87\%, showing that the label prefix brings only marginal improvement and has a decrease of 0.22\% compared to Ours.
Moreover, case (b) outperforms traditional caption-only case (c) and label-only case (d) by 8.77\% and 5.76\%, respectively, confirming that high-quality reasoning texts are key to ReAlign’s superior generalization and detection performance.

\noindent  \textbf{Ablation Study on Training Configurations.} In ReAlign, we fine-tune the image encoder using LoRA, optimizing with a combination of classification loss and contrastive loss. We aim to compare the impact of joint optimization versus sequential optimization for the alignment and classification. We also examine the impact of different image encoder fine-tuning methods (full parameter, LoRA, and freezing). We designed five variants and evaluated these settings on the UltraSynth-10k, as shown in Tab.~\ref{table:ablation2}. 
Comparing ours with case (c), sequential optimization is 13.01\% lower than joint optimization, indicating that, under classification constraints, the image encoder is better able to efficiently learn AIGI detection-related information from textual representations.
Comparing ours with case (a), where the image encoder is fully fine-tuned, we observe a 2.4\% drop to 94.69\%, highlighting that LoRA fine-tuning preserves general semantics perception while enhancing detection of fake samples.
Case (b) further confirms the above conclusions.
In case (d) and case (e), we freeze or use LoRA to fine-tune the image encoder and optimize the model using only the classification loss, achieving 89.15\% and 93.68\% respectively, thus demonstrating the effectiveness of our reasoning-alignment mechanism. More results are presented in \textcolor{blue}{\textbf{\textit{supplementary materials}}}.

\begin{table}[t]
\caption{\textbf{Accuracy (\%) under Different Training Configurations.} ``Joint'' and ``Sequential'' denote joint and sequential optimization of the alignment and classification tasks, respectively.}
\begin{center}
\vspace{-20pt}
\resizebox{1.0\linewidth}{!}{
\begin{tabular}{c|cc|ccc|c}
\toprule
\multirow{2}{*}{\textbf{Case}} & \multicolumn{2}{c|}{\textbf{Training Strategy}} & \multicolumn{3}{c|}{\textbf{Fine-tuning Method}} & \multirow{2}{*}{\textbf{Mean Accuracy}} \\
\cmidrule(lr){2-3} \cmidrule(lr){4-6}
\textbf{} & \textbf{Joint} & \textbf{Sequential} & \textbf{Full} & \textbf{LoRA} & \textbf{Freeze} & \textbf{} \\
\hline
Ours & \checkmark & \textbf{} & \textbf{} & \checkmark & \textbf{} & 97.09 \\
\hline
(a) & \checkmark & \textbf{} & \checkmark & \textbf{} & \textbf{} & 94.69  \textcolor{DarkGreen}{(-2.40\%)} \\
(b) & \textbf{} & \checkmark & \checkmark & \textbf{} & \textbf{} & 79.07  \textcolor{DarkGreen}{(-18.02\%)} \\
(c) & \textbf{} & \checkmark & \textbf{} & \checkmark & \textbf{} & 84.08  \textcolor{DarkGreen}{(-13.01\%)} \\
(d) & \textbf{} & \textbf{} & \textbf{} & \textbf{} & \checkmark & 89.15  \textcolor{DarkGreen}{(-7.94\%)} \\
(e) & \textbf{} & \textbf{} & \textbf{} & \checkmark &  &  93.68 \textcolor{DarkGreen}{(-3.41\%)} \\
\bottomrule
\end{tabular}
}
\end{center}
\label{table:ablation2}
\vspace{-20pt}
\end{table}
\section{Conclusion}

In this work, we examine the strengths and weaknesses of both LLM-based and non-LLM AIGI detection methods and highlight the importance of reasoning texts generated by RL-optimized LLMs. Our analysis shows that their effectiveness largely stems from three properties: strong discrimination, cross-domain generalization, and sensitivity to semantic errors. Building on these insights, we introduce ReAlign, a lightweight framework that aligns visual features with reasoning-based textual representations through contrastive learning. Trained with a joint alignment and detection objective, ReAlign achieves SOTA performance and strong generalization across multiple benchmarks, including the new UltraSynth-10k dataset. By combining the semantic richness of reasoning-based representations with the efficiency of a compact detector, ReAlign effectively unifies the advantages of both paradigms.

Looking ahead, ReAlign provides a promising foundation for scalable and cost-efficient AIGI detection systems, paving the way toward stronger information-security defenses and a more robust AI safety infrastructure.

{
    \small
    \bibliographystyle{ieeenat_fullname}
    \bibliography{main}
}


\end{document}